\documentclass{IOS-Book-Article}

\usepackage{times} % assumes new font selection scheme installed
\usepackage{amsmath} % assumes amsmath package installed
\usepackage{amssymb}  % assumes amsmath package installed
\usepackage{color}
\usepackage{bm}
\usepackage{caption}
\usepackage{afterpage}
\usepackage{array,graphicx}
\usepackage{wrapfig}
\usepackage{amsfonts}
\usepackage{multicol}
\usepackage[titlenumbered,ruled,linesnumbered,algoruled,boxed,lined,algo2e]{algorithm2e}
\usepackage[noend]{algpseudocode}
\def\hb{\hbox to 10.7 cm{}}

\begin{document}

\pagestyle{headings}
\def\thepage{}

\begin{frontmatter}              % The preamble begins here.

%\pretitle{Pretitle}
\title{Visual search and recognition for robot task execution and monitoring}

%\markboth{}{March 2018\hb}
%\subtitle{Subtitle}
\author[1]{\fnms{Lorenzo} \snm{Mauro}},
\author[1]{\fnms{Francesco} \snm{Puja}},
\author[1]{\fnms{Simone} \snm{Grazioso}},
\author[1]{\fnms{Valsamis} \snm{Ntouskos}},
\author[1]{\fnms{Marta} \snm{Sanzari}}, 
\author[1]{\fnms{Edoardo} \snm{Alati}},
\author[1]{\fnms{Fiora} \snm{Pirri}}

\address[1]{ALCOR Lab, DIAG, Sapienza University of Rome, Italy} 
{\tt\small \{mauro, puja, grazioso, ntouskos, sanzari, alati, pirri\}@diag.uniroma1.it}\\

\begin{abstract}
Visual search of relevant targets in the environment is a crucial robot skill.  We propose  a preliminary framework for the execution monitor of  a robot task, taking care of the robot attitude to visually searching the environment for targets involved in the task.   Visual search is also relevant to recover from a failure.
  The framework exploits deep reinforcement learning  to acquire a {\em common sense} scene structure and it takes advantage of a deep convolutional network to detect objects and relevant relations holding between them. The framework builds on these methods to introduce a vision-based execution monitoring, which  uses classical planning as a backbone for task execution.
  Experiments show that with the proposed vision-based execution monitor the robot can complete simple tasks and can recover from failures in autonomy. 
\end{abstract}

\end{frontmatter}

\section{Introduction}
One of the main problems in robot task execution is the interplay between perception and execution. Indeed, it is not clear what information a robot should collect from the environment to be able to continuously interact with it.

\begin{figure}[!htb]
    \centering
    \includegraphics[width=0.98\linewidth]{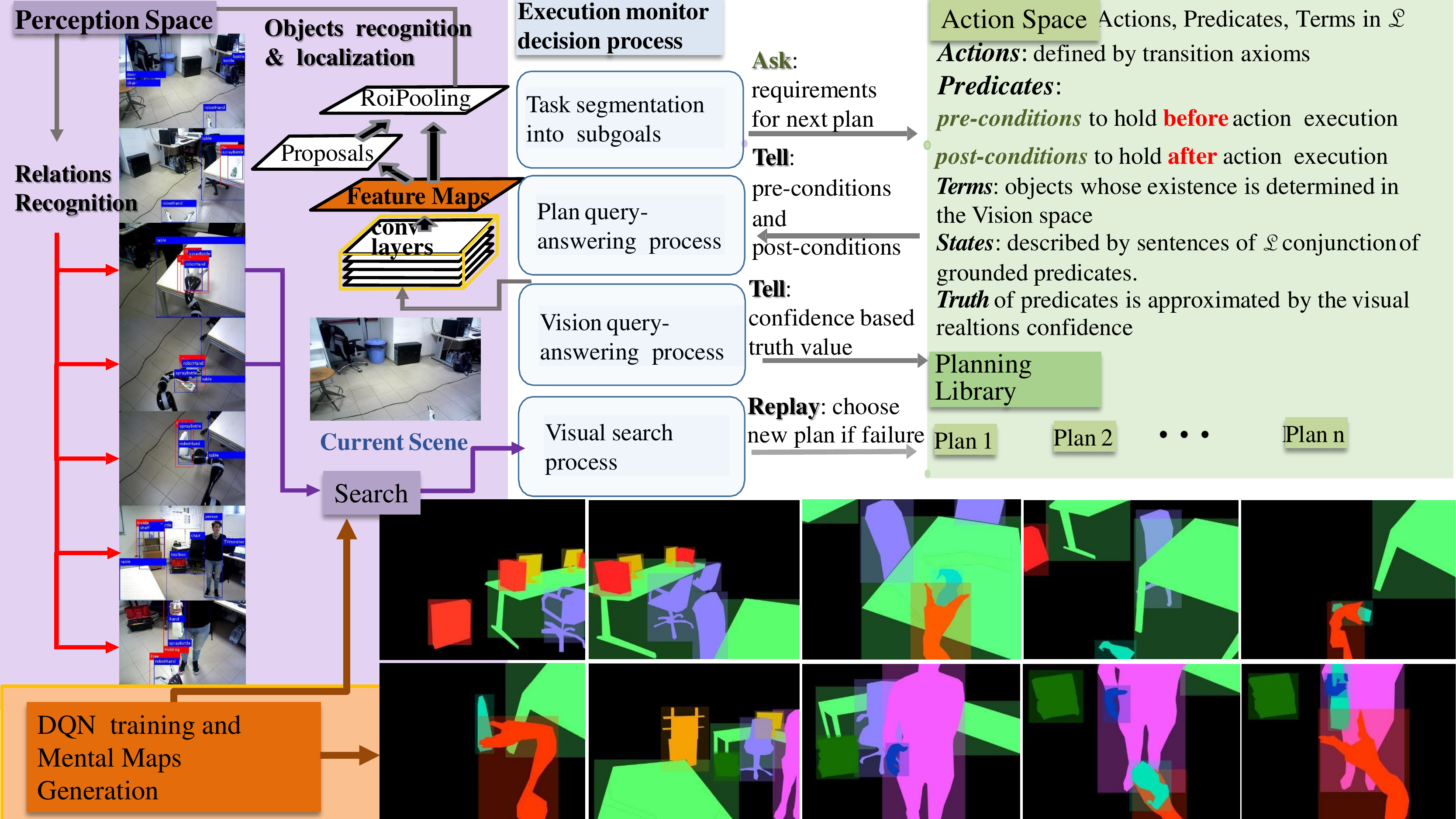}
    \caption{The execution monitor framework combining classical planning, detection and localization of objects together with visual search. Best seen on screen.}
    \label{fig:framework}
\end{figure}

 So far, 3D reconstruction, 3D reconstructed scene segmentation and SLAM seemed the best way to provide an effective and convenient representation of the environment for task execution. Several scientific contributions have focused on solving these problems in compelling real-world scenes. Once the system is aware of both the complete environment and the robot pose, it provides to the execution monitor adequate instructions for both the planning and  the required continuous updates. 

A major problem in 3D reconstruction and SLAM approaches is that they do not consider the robot gaze direction. They create scene reconstructions designed for the human user and not in order to facilitate the robot reasoning about the environment. Indeed, this lack of connection between the scene and where the robot looks at creates an unsound reasoning process: the map guides the robot searching according to a global view (the human user view) and not on a saliency-based view, \cite{Ntouskos-2013BICA,qodseya2016eccvw}.  

As a matter of fact, most of a 3D map reconstructions are full of unnecessary details for a  robot to keep in memory, in relation to its job. It has also to be said that a representation of the world needs to cope with frame and qualification problems \cite{mccarthy1981, pirri1999} and when it comes to a full 3D map a huge amount of details  can be very  hard to deal with.  In fact, humans learn to  search and  move in any complex environment essentially relying on foveated vision and on what they have learned, to limit waste of their own resources.

Basing on the human ability to hold partial representations and to fill the gap between what the robot can denote and the complexity of the world, we introduce a parsimonious representation, which we call {\em mental map}. Mental maps are scene snapshots keeping only objects and relations that the robot can detect and supporting the learning process for the visual search: they are the robot cartoon representation of its environment. Learning where to look at solves the problems related to both global and local maps,  as it can help  to assess a low cost, yet effective,  representation of the environment.

Here we propose a novel perspective to execution monitor, which is based on the interplay between execution and perception. Our idea (see Figure \ref{fig:framework} for an overview) combines deep learning, which is able to develop a quite mature cognitive approach to perception, with classical planning and monitoring. 
%%%
The proposed framework is centered on the robot execution monitor. The execution monitor is supported by the planner, see \cite{mcdermott-1998, helmert-2009, cashmore2015}, by the vision system processing visual data with Faster R-CNN \cite{ren-2015}, and by the learned ability to direct the robot gaze in a plausible direction for searching the environment, via a modified version of DQN \cite{mnih2015}. Planning is the backbone, in the sense that it provides the properties of the world that are expected to be true before and after the execution of a robot action. The planner, however, does not decide actions sequences since it cannot verify whether the prescribed requirements are true. The planner provides a sort of prior knowledge about the space of action execution, but not of their realizations. That is why the planner is enriched by a query-answering process with the execution monitor to validate whether the requirements are satisfied in the environment.  The execution monitor resorts to visual search to activate the choice of next plan, which is crucial in case of failure (e.g. the loss of an object). To infer policies the execution monitor  builds states representations of the scene out of the visual stream results,  formed by mental maps, see Figure \ref{fig:framework}. Mental maps incorporate the essential knowledge for a robot to complete a task. To learn the search ability, we adapted the Deep Q-Network, \cite{mnih2015}, to receive RGB input and we train it on a virtual environment.
Policies learned on the virtual environments, though displaying a good ability to deal with the robot representation via mental maps, can be improved by learning in real environments, and this is the objective of future work. The execution monitor is  finally able to control the loop between action and perception without the need to keep the structure of the entire environment in memory, but updating its perception with what is required/asked by the planner to complete the action.

\begin{figure}[t!]
 \centering
  \includegraphics[width=.99\textwidth]{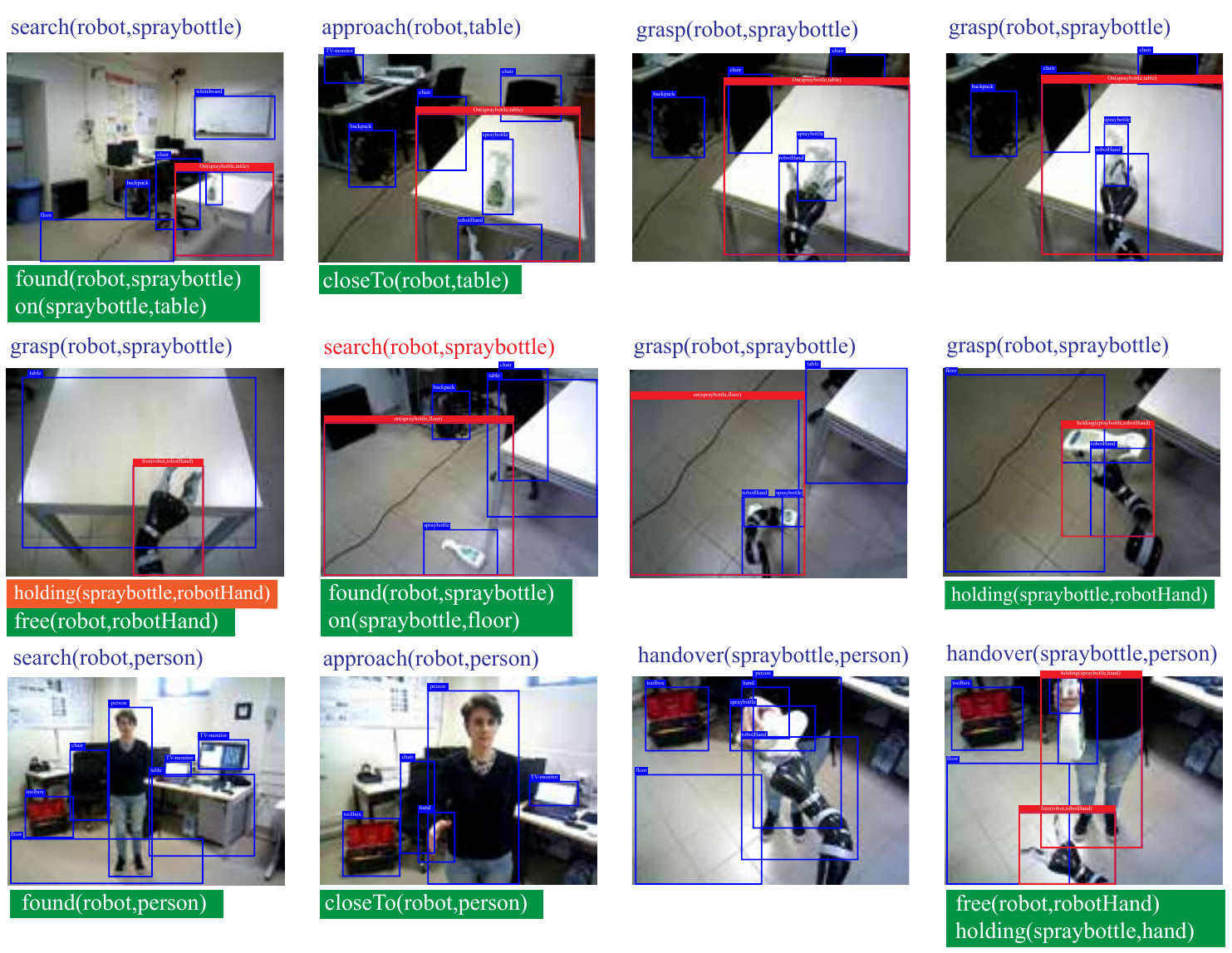}\\
 \caption{Execution monitoring  of the task {\em take the spray bottle and hand it to Person}. Actions written in blue are provided by the planner and those in red by the visual search policy, once a failure is detected. The post-conditions are shown below the images, in orange if not verified. Inside the images, the bounding boxes of the detected objects are shown in blue and in red the detected relations calculated by the perception system.}\label{fig:overview}
\end{figure}

The robot operates in a laboratory environment and a schematic overview of a task execution is given in Figure \ref{fig:overview}.

\label{sec:intro}

\typeout{********************** RELATED *****************************************}
\section{Related Work}
\paragraph{\textbf{Visual Search}}
Visual search is not a novel topic in robotics and a  large body of work has been done so far to approach the problem, though it has  been faced mainly in terms of global navigation.

Some of the approaches  based on map navigation require a prior knowledge of the environment to make decisions  (see for example \cite{borenstein1989real, borenstein1991vector, kim1999symbolic, oriolo1995line}), all based on 3D map reconstruction, as in \cite{whitaker1998,kruijff2016deployment}.  
 Other  methods do not actually rely on pre-saved maps but  on a map reconstruction which is done either on the fly (see \cite{sim2006autonomous, wooden2006guide, davison2003real, tomono20063}) or during an initial human guided training phase as in (\cite{kidono2002autonomous, royer2005outdoor}). Other approaches instead face the search in terms of obstacle avoidance using input images in so not relying  on any kind of map (see, \cite{haddad1998reactive, lenser2003visual, remazeilles2004robot, saeedi2006vision}).
However, all of the above-mentioned methods depend in one way or another on the navigation of a reconstructed environment. 
To be able to properly explore the environment in this way, the process needs to store the collected information in memory. 

Differently from previous works our approach aims at updating the perception with only what is necessary at the very moment in which the information is needed to the robot to perform and complete what it is doing. This approach is more similar to a human-like behavior and it is realized with a neural network trained with reinforcement learning. 

Earlier approaches to reinforcement learning (RL) (see, for example, \cite{bertsekas1995,sutton1998} made the effort to extend this technique  to a very large set of actions and to make it a robust method. Despite that, the breakthrough  started with the work described in \cite{mnih2015} who introduced a new amazing research area, the deep RL \cite{mnih2016,levine2016}. The main difficulty with this approach is the need to simulate the environment to perform the required huge collection of  experiments. Hence a certain amount of research has been devoted creating new simulation environments such as \cite{mottaghi2016, zhu2017}. Here we contribute with the above mentioned mental map representation of the environment to be distinguished from the V-REP \cite{rohmer2013} simulation used for training. A similar approach  has been proposed by \cite{zhu2017}. Differently from their work, which uses an auxiliary target image as input to the network, here we only provide the robot's onboard camera pictures as an input to the visual search, allowing us to search for objects regardless of their orientation.

\paragraph{\textbf{Execution Monitoring}}

%%%
Managing  the execution of actions  involve facing several difficult challenges.
A great number of researchers tackled the non-deterministic response of the environment to robot actions for high-level robot tasks \cite{pettersson2005, ingrand2017}. Some introduced the role of perception in execution monitoring \cite{doyle1986}, others focused on the recovery from errors that could occur at execution time \cite{wilkins1985} and on relations recognition,  which recently raised much interest in computer vision community \cite{guadarrama2013, Lu-2016, das2016} though related to 2D images. However, many problems remain to be addressed and there is still a lack of a comprehensive approach to address all the problems concerning the interleaving of perception and actions in a coherent way.

The first definitions of execution monitoring in non-deterministic environments have been presented in \cite{fikes1971, nilsson1973}. Recovery from  errors that could occur at execution time was already faced by \cite{wilkins1985} and equally the role of perception in execution monitoring was already anticipated in the work of \cite{doyle1986}. Due to the difficulties in dealing with scene recognition the efforts have been directed toward models, such as in \cite{sutton1998, bertsekas1995}, that manages the effects of actions  allowing the execution in partially observable environments \cite{boutilier2000}. In high level monitor several authors, among which  \cite{hornung2014,mendoza2015}, have been recently addressing the integration of observations. Still, the most important development has been achieved with deep learning for perception and in particular with DCNN\cite{guadarrama2013},  deep Reinforcement Learning (RL) \cite{mnih2015,mnih2016}  and visual planning \cite{ZhuGKFFGMF17}. 

Nevertheless, despite the great number of contributions \cite{guadarrama2013,Lu-2016,das2016} for 2D images, an appropriate relations recognition  method for robot execution is still not available. For this reason, we exploit the robot RGBD vision and a simple rough segmentation method within the bounding box delivered by Faster R-CNN to obtain a more detailed object region, in so ensuring a better approximation of relations.

\typeout{********************** EXEC MON *****************************************}

\section{The Execution Monitor}
The robot execution monitor drives the interplay between the action space, defined  by actions, relations, and objects in a first order typed language and the perception space that specifies objects and relations in terms of features and measures.   Figure \ref{fig:framework} provides a schematic representation of the interplay between the two spaces.

The action space addresses the current knowledge required by the robot in order to perform a specific action, whether this is an action not affecting the world, or  an action changing the state of the world. The execution monitor, given a goal in terms of a task, has a strategy to segment it into sub-goals. This strategy is not described here. For each sub-goal, according to the terms appearing in it, the execution monitor uses naive Bayes to {\em ask} to the plan library the plan matching the sub-goal,  according to the current state. The current state is verified by assessing what is available to the robot perception in terms of objects positions and relations holding at the specific time.  Plans are defined in the classical PDDL framework \cite{mcdermott-1998, helmert-2006}, see Section \ref{sec:detPlanner}. The inference of a sub-goal by the chosen plan results in an ordered list of actions, objects, and relations that the planner {\em tells} to the execution monitor. The execution monitor,  in turn, asks  the vision process to verify the list of actions while they are executed. If the list is verified the execution monitor triggers the next sub-goal. Otherwise, this component activates the visual search to focus on the validation of the plan requirements, and eventually repeats the current plan from the specific state. An example of a failure is given in Figure \ref{fig:overview} where in sub figure 5 the robot hand accidentally drops the spray bottle.\label{sec:execmon}

\typeout{********************** Perc Syst *****************************************}
\section{The Perception system}
The perception space is defined by features extracted from the DCNN and by the mental maps. Two processes live in this space: the object detection and localization and the visual search.

\subsection{Object detection and localization}\label{sec:visualSystem}
\typeout{********************** Vis System *****************************************}
Object detection uses the well-known Faster R-CNN deep convolutional neural networks  \cite{ren-2015} and a statistic estimation of the object boundary within the bounding box detected by the trained DCNN, using the active features of the VGG-16 5.3 layer, which is a component of Faster R-CNN.

The combination of segmented depth and object labels allows inferring visual relations from the robot point of view. 

Specifically, to detect both objects and relations we have trained Faster R-CNN \cite{ren-2015} on ImageNet \cite{krizhevsky2012}, Pascal-VOC \cite{everingham2015} dataset, and on images taken on site. 
We have trained 5 models to increase the accuracy of the detection obtaining good results. The detection accuracy is based on a confidence value measured on a batch of 10 images, taken at $30fps$, computing the most common value in the batch and returning the sampling mean accuracy for that object.

For each object $ob_j$ detected by the DCNN we estimate a statistics of the active features with dimension $38\times 50\times 512$, taken before the last (5.3) convolution layer, at each pixel inside the recognized object bounding box $bb(ob_j)$. With this features set we  estimate the probability that features with the highest frequency in the feature volume, fall within the object. Then reprojecting the location of maximum probability on the object cropped by the bounding boxes we obtain regions within the object. It is interesting to note, indeed, that the features  (the weights learned by the convolution network) of the 5.3 convolution layer of the VGG-16, behaves like saliency in identifying the object interesting parts. 
Then, using nearest neighbor we find  more accurate regions within the object. Since the images/videos available to robot perception are RGBD images, from this process we obtain a quite accurate segmentation of the depth map of the object.
Note that this step is necessary to recognize relations between objects since we are talking about spatial relations where distance and location are crucial. Considering only the bounding boxes would hinder spatial relations: some of the objects involved in the robot tasks, such as a table, ladder, spray bottle and so on, have bounding boxes containing also other parts of the environment (see Figure \ref{fig:overview}).
Without an accurate segmentation, the detection of an object returns incoherent depth and the decision for execution monitoring on controlling actions, such as \emph{grasp}, \emph{hand{-}over}, \emph{place}, \emph{open}, \emph{close} and so on, cannot take place. Instead with the segmentation easily obtained via the significance of the last convolutional layer of VGG-16, we are able to define spatial relations such as {\em Holding, On, Below} and so on, just using relations derived by the center of mass of the objects, in close connection to  the qualitative spatial relations framework \cite{cohn1997} extended to 3D, see also \cite{kunze2014}. A set of relations used in the framework are shown in Table \ref{tab:ablat} evaluating accuracy and ablation study. Evaluation of objects detection and localization is reported in the result Section \ref{sec:results}.

\begin{table}
\centering
%\resizebox{0.99\columnwidth}{!}{
\begin{tabular}{|l|c|c|c|c|c|c|}
\hline
\textbf{Predicate} & \textbf{Full} & \textbf{BB} & \textbf{GT masks} & \textbf{no prior} & \textbf{no shape} & \textbf{no depth feat.} \\[2mm]
CloseTo  & 82\% & 79\% & 89\% & 79\% & 72\% & 49\% \\
Found  & 88\% & 85\% & 95\% & 85\% & 80\% & 61\% \\
Free & 91\% & 86\% & 91\% & 86\% & 83\% & 68\% \\
Holding & 79\% & 72\% & 88\% & 75\% & 74\% & 56\% \\
Inside & 69\% & 64\% & 87\% & 71\% & 65\% & 57\% \\
On & 85\% & 77\% & 96\% & 79\% & 78\% & 65\% \\
InFront & 88\% & 81\% & 95\% & 84\% & 83\% & 63\% \\
Left & 90\% & 81\% & 95\% & 85\% & 86\% & 72\% \\
Right & 83\% & 79\% & 91\% & 79\% & 80\% & 61\% \\
Under & 80\% & 76\% & 88\% & 79\% & 76\% & 59\% \\
Behind & 85\% & 78\% & 81\% & 76\% & 79\% & 61\% \\
Clear & 82\% & 75\% & 80\% & 73\% & 73\% & 60\% \\
Empty & 83\% & 72\% & 78\% & 79\% & 68\% & 61\% \\
\hline
\textbf{Average} & \textbf{84\%} & \textbf{78\%} & \textbf{92\%} & \textbf{80\%} & \textbf{78\%} & \textbf{61\%}\\
\hline
\end{tabular}
%}
\caption{Accuracy and ablation study of predicate grounding. \textbf{Legend:}  \textit{BB}: bounding boxes, \textit{GT masks}: ground truth segmentations masks, \textit{no prior}: without use of VGG-features \textit{no shape}: without use of statistics. \textit{no depth feat.}: without use of depth features.}\label{tab:ablat} 
\end{table}

\subsection{Visual Search}\label{sec:visualSearch}
\typeout{********************** Vis Search *****************************************}
\begin{algorithm2e}[ht!]
\SetKwInOut{Input}{input}
\SetKwInOut{Output}{output}

\Input{Faster R-CNN network \newline Simulated Environment}
\Output{Trained DQN network}
 Load Simulated environment \;
 Load Faster R-CNN network \;
 Initialize replay memory $D$\;
 Initialize action-value function $Q$ with random weights $\omega$\;
 Initialize first state $s_1$ and Mental Map $\theta_1$ with random values\;
 \For{t $= 1,maxSteps$}{
    Calculate $\varepsilon$-greedy probability $\varepsilon(t)=0.9/t_{\varepsilon_{min}}*t+1$\;
 	With probability $\varepsilon(t)$ select a random action \newline $a_t$ 	otherwise select $a_t= \underset{a}\max Q(\theta(s_{t}),\omega)$\;
 	Execute action $a_t$ and observe image $i_{t+1}$\;
 	Elaborate image $i_{t+1}$ with Faster R-CNN \;
 	Produce processed picture $i_{t+1}^*$ and reward $r_t$\;
 	Set state $s_{t+1}=s_t, a_t,i_{t+1}^*$\;
 	Create mental map $\theta_{t+1}=\theta(s_{t+1})$\;
 	Store transition $(\theta_{t}, a_t, r_t, \theta_{t+1})$\;
  	\If{t $>startTraining$}{
  		Sample random minibatch of transition  $(\theta_{j}, a_j, r_j, \theta_{j+1})$ from $D$\;
  		Set
		\[y_j =
 			\begin{cases} 
				r_j  & \mbox{for terminal } \theta_{j+1} \\
				r_j + \gamma \underset{a}\max Q_\pi(\theta_{j+1},\omega) & \mbox{for non terminal } \theta_{j+1}
 			\end{cases}
		\]
		Perform gradient descent on loss $(y_i - Q(\theta_{j},\omega))^2$
   	}
 }
 \caption{Deep Q-Learning with Mental Maps and Experience Replay}\label{alg:DRL}
\end{algorithm2e}
%}
%\end{minipage}

As introduced above, {\em visual search} is triggered both to check the current state and to recover from a failure. Visual search  is modeled by deep reinforcement learning  to focus the robot toward the items of interest. This step is necessary for the robot to find out whether the preconditions of the action to be executed are satisfied or to search a starting point for a recovery action. 

The reinforcement learning method allows learning by trial and error how to choose the best action to locate the target object. The visual search does not have to rely on previous knowledge of the environment but it only uses the images returned by the onboard robot camera suitably elaborated into mental maps.
To achieve this result, the Deep Q-Network (DQN) proposed in \cite{mnih2015} is used, adapted to receive as input a collection of preprocessed RGB-D camera frames called \emph{Mental Maps}. We recall that the target of the DQN learning process is to learn an approximation of the reward function for each possible action: the Q-value. The final policy will be, therefore, the one created by selecting the action with the highest Q prediction.

Since the field of vision of the agent returns only a partial observation of the environment, the states are only partially observable. To treat this setting as a fully observable one, the current state is composed of a history of the previous frames. However, due to memory limitations, only the last $15$ states are used. Together they form the \emph{Mental Map}, which is the input of the DQN.

The raw frames that compose the \emph{Mental Map} are preprocessed with Faster R-CNN as explained in Section \ref{sec:visualSystem}. Using the output of the network we draw a colored mask for each detected object. The color is assigned in relation to the object class. A visualization of the framework can be found in Figure \ref{fig:VisSearchframework}.

The creation of the mental maps is done for a variety of reasons. The strong visual clue introduced by the segmentation of the objects detach the learned behavior from the specific textures of the images. This property makes it possible to use the same DRL model in scenarios which have the same objects but different textures by just fine tuning the object detection network. In this way, it is possible to use a robot trained in a simulated environment directly in a real-world application.
Strong visual clues allow also to analyze the situation regardless of the specific object orientation. This is a crucial step, in particular during the failure recovery scenario when the object orientation and position are unknown. Without this step, the DRL method would need to be trained not only on a wide variety of target object positions but also on a wide variety of target object orientations making the training process longer.

The mental maps generation process produces the reward used to calculate the training loss. This reward is calculated in relation to whether or not the robot has found the target object. In those steps in which the target is recognized a positive reward is assigned in proportion to the classification score of Faster R-CNN. Instead, the reward of any action not leading to the localization of the target is set to a fixed negative value.

At training time the policy used is $\epsilon$-greedy. The value $\epsilon$, which is the probability that the training process chooses a random action instead of the one calculated by the network, starts from an $\epsilon_{max}$ value and linearly descends to an $\epsilon_{min}$ value. In this way, the agent is able to explore the environment and to focus on optimizing the policy as the training goes on.
An important feature of our training process is that the episode has a fixed step size. The sequence is not interrupted even when the robot sees the target object with a confidence score higher than the decision threshold. This is done in order to choose the action that maximizes the classification score and the consequent reward. In this way, if the robot sees the target object classified with a low score, it learns to investigate on the situation  trying to increase the classification score by getting closer or looking at the object from a different position. Furthermore, with such investigating approach, if the classification score is the result of a false positive, the score would drop down prompting the robot to change the search behavior. Algorithm \ref{alg:DRL} illustrates the main steps of the training process.

\label{sec:perception}

\typeout{********************** ACtion Space *****************************************}
\section{Action space and the plan library}\label{sec:detPlanner}
We focus here on the deterministic part of the visual execution space: the action space and the plan library. The action space is defined by  a {\em domain} ${\mathcal D}$  formed by objects, predicates and real numbers,  which are represented in the {\em robot language} ${\mathcal L}$.   A term of ${\mathcal L}$ can be an object $ob_i$ of the domain ${\mathcal D}$, a variable or a function $f:{\mathcal D}^n\mapsto {\mathcal D}$, such as actions. Predicates are the counterpart of relations in the perception space and are here assumed to be only binary. Predicates are defined exclusively for the specified set of objects of the domain and are interpreted essentially as spatial relations such as {\em On, CloseTo, Inside, Holding}.

A {\em robot task} is defined by a list of plans $\langle {\mathcal P}_1, \ldots, {\mathcal P}_m\rangle$; a {\em plan} ${\mathcal P}_i$ is defined as usual  by an initial state and a set of rules (axioms) specifying the preconditions $H(a_{t})$ and postconditions $K(a_t)$  of  action $a(t)$, where $t\in {\mathbb N}$ is a discrete time index. To simplify this presentation we assume that preconditions and postconditions are  conjunctions of ground atoms (binary relations), hence a state $s_t {=} H(a_t) \cup K(a_t)$.

In our settings robot {\em actions} can be of two types: (1) {\em transforming actions}, which are  actions changing the state of the world, affecting some object in the world, such as {\em grasp}, {\em hand{-}over}, {\em place}; (2) {\em ecological actions}, which are actions changing the state of the robot, such as {\em move-Forward},  {\em approach} $ob_j$  along a direction, or {\em look-Up}, {\em look-Down}.

According to these definitions, a {\em robot plan} specifies at most one transformation-action.  This action is a {\em durable action},  defined by an action {\em start} and by an action {\em end}. Any other action in the plan is an ecological action.  In particular, given the initial state, each plan introduces at most a new  primal object  the robot can deal with and the relations of this object with the domain. An example is as follows: the robot is holding an object, as a result of a previous plan, and it has to put it away on a toolbox. This ensures that recovery from a failure is circumscribed for each plan to a single transformation action.

The {\em task goal} is unpacked into an ordered list of goals ${\mathcal G}{=}(G_1,\ldots, G_m)$ such that each $G_i$ is the goal of plan ${\mathcal P}_i$ in the list of plans forming a task. 
Given a goal $G_i$, if a  plan ${\mathcal P}_i$ leading from the initial state to the goal exists, then a sequence of actions leading to the goal is inferred by the search engine \cite{helmert-2006}.  Here plans are defined  in PDDL \cite{edelkamp-2004}. 

We extend the parser introduced in \cite{helmert-2006} to infer plans so as to obtain the initial state, the sequence of actions $(a_0, \ldots, a_n)$ leading to the goal, and the set  of states $S_{{\mathcal P}_i}$ for each plan ${\mathcal P}_i$. This list  forms the set of deterministic states and transitions.

\typeout{********************** Results *****************************************}
\section{Results}\label{sec:results}
\noindent
\subsection{Experiments setup}
The framework has been tested under  different conditions,  in order to evaluate different aspects of the model. To begin with, all experiments have been performed with a custom-made robot.  A Pioneer 3 DX differential-drive robot is used as a compact mobile base. To interact with the environment we mounted a Kinova Jaco 2 arm with 6 degrees of freedom and a reach of 900mm and finally, for visual perception, we used an Asus Xtion PRO live RGB-D camera mounted on a Direct Perception PTU-46-17.5 pan-tilt unit. 
We ran the experiments on two computers mounted on the robot.  One was dedicated to the planning and management of the execution and another one, equipped with a Nvidia GPU, to run the perception system.

For training and testing our system, we used different objects and all the relations needed to correctly execute and monitor different tasks. A list of those elements can be seen in Table \ref{tab:relations}. 

%%%%%%%%%%%%%%%% RELATIONS & OBJECTS %%%%%%%%%%%%%%%%%%
\begin{table}[t!]
\caption{Subset of Objects, relations, transformation actions and ecological actions of the robot language ${\mathcal L}$}\label{tab:relations}
\centering
%\resizebox{0.99\columnwidth}{!}{
\begin{tabular}{l|l|l|l}
\textbf{\scriptsize Relations} & \textbf{\scriptsize Objects} & \textbf{\scriptsize Transformation actions} & \textbf{\scriptsize Ecological actions} \\ \hline
\scriptsize CloseTo & \scriptsize Brush & \scriptsize Close & \scriptsize Look-down \\%[-0.1ex]
\scriptsize Found & \scriptsize Chair & \scriptsize Grasp & \scriptsize Look-left \\%[-0.1ex]
\scriptsize Free & \scriptsize Cup & \scriptsize Open & \scriptsize Look-right \\%[-0.1ex]
\scriptsize Holding & \scriptsize Floor & \scriptsize Hand-over & \scriptsize Look-up\\%[-0.1ex]
\scriptsize Inside & \scriptsize Hammer & \scriptsize Place &  \scriptsize Move-forward \\%[-0.1ex]
\scriptsize On & \scriptsize Person & \scriptsize Lift &  \scriptsize Move-backward\\%[-0.1ex]
\scriptsize InFront& \scriptsize Spray Bottle & \scriptsize Push  & \scriptsize Turn-left\\%[-0.1ex]
\scriptsize Left & \scriptsize Screwdriver &\scriptsize Spin & \scriptsize Turn-right\\%[-0.1ex]
\scriptsize Right & \scriptsize Shelf & Dispose  & \scriptsize Localize \\%[-0.1ex]
\scriptsize Under & \scriptsize Toolbox  & & \scriptsize Rise-arm \\%[-0.1ex]
\scriptsize Behind & \scriptsize TV-Monitor & &   \scriptsize Lower-arm \\%[-0.1ex]
\scriptsize Clear & \scriptsize Table &  & \scriptsize Close-Hand\\
\scriptsize Empty & \scriptsize Door & & \scriptsize Open-Hand\\
\end{tabular}
%}
\end{table}

\subsection{Results for Object detection, localization and relations recognition}

We trained the visual system with images taken from ImageNet and Pascal VOC datasets as well as images collected with the RGB-D camera of the robot, and using predicates for the execution of the tasks. To train the DCNN models we divided the images into training and validation sets with a proportion of 80\%-20\%. We have chosen five different models for the objects, performing 70000 training iterations for each model on a PC equipped with 2 GPUs.
The visual system has been tested under different conditions, in standalone tests and during the execution of different tasks. The accuracy has been computed considering the batch of 10 images, see in Section \ref{tab:relations}.

\typeout{************************ HERE}

\subsection{Results for Visual Search}
\begin{figure}[!htb]
    \centering
    \includegraphics[width=0.99\linewidth]{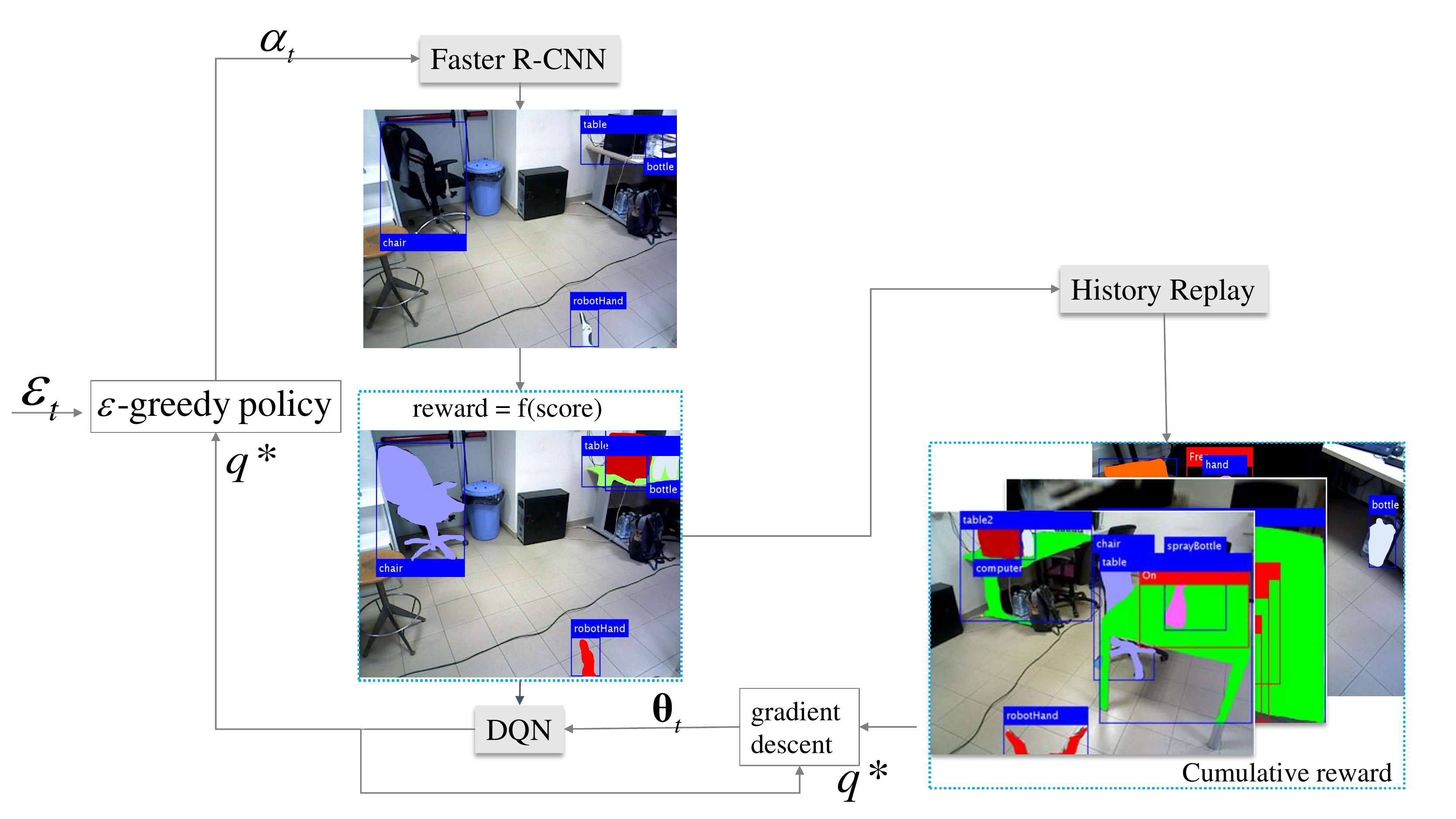}
    \caption{Visual Search Training Framework. In the figure $\alpha_t$ is the action chosen by the $\epsilon$-greedy policy at time $t$, $q^*$ is the q-value calculated by the DQN network and ${\bm \theta_t}$ is the vector of  network weights at time $t$.}
    \label{fig:VisSearchframework}
\end{figure}

\paragraph{\textbf{Simulation in V-REP:}}
The training phase of the \textit{visual search} component requires a huge amount of steps and episodes in order to perform well at testing time. Training the robot in a real environment is not an efficient option.
To address this problem we created a virtual environment using the V-REP simulation software \cite{rohmer2013v}, that allows training the robot in a setting similar to a real scenario. The simulation only generates camera images at the end of the chosen action since this is the only information available to a robot in a real scenario. In this way, the training process can be easily extended also for a real robot and the learned policy can be fine-tuned in a real scenario after a long training in a simulated environment.
The robot can perform several actions, it can move backward or forward, turn right or left and it can also move the camera upward or downward and to the right or to the left.
The policy chooses the action and when it is performed, the image from the real or simulated camera is given to the Mental Maps creation process as shown in Figure \ref{fig:VisSearchframework}.

\paragraph{\textbf{Training Details:}}
To train the robot in a wide variety of situations, the target object, camera position and orientation are changed at every episode. Further, before starting an episode, a random number (between 1 and 3) of random actions are performed. In this way, we have 46720 different combinations of initial settings to prevent the overfitting problem for training.

The Deep reinforcement learning is implemented with a modified version of the DQN implementation, see \cite{mnih2015}. The network is adapted to take as input the last $15$ RGB-D frames with a resolution of $640 \times 480$ pixels. The network has 3 convolutional layers each followed by a ReLu layer. The first one has $15 \times 4 \times 4 \times 4 \times 8$ filters and $1 \times 4 \times 2 \times 2 \times 1$ stride. The second layer has $1 \times 8 \times 8 \times 8 \times 64$ filters and  $1 \times 1 \times 4 \times 4 \times 1$ stride and the third and final layer has  $1 \times 4 \times 4 \times 64 \times 128$ filters and  $1 \times 1 \times 2 \times 2 \times 1$ stride. The returning feature map from the convolutional layers is reshaped and passed to $2$ fully connected layers.
The performances reached are the results of more than 4 million iterations on a machine with 16 Gb of RAM, processor Intel Core i7-3770 and graphics card Geforce GTX 1080. The first 160 steps are used to fill enough the replay memory to allow a correct sampling of 32 steps sized mini-batches. The maximum replay memory size is set to 1000 steps and the weight update is done every 4 steps. The mental maps length is set to $15$ steps and the $\epsilon$ is linearly decreased from 1 to 0.1 over the first half of the training steps.
Different episode settings have been tested, changing essentially the condition to stop the episode and the reward function. The final solution is to assign a positive reward to frames in which the object is detected with both a high or low classification score, proportionally to the magnitude of the score itself. The final reward function is set to $R(score) = 50*score^5$ for scores higher than 0.3 and -0.1 for smaller scores. The episode length is set to a fixed size of 50 steps.

\begin{figure}[!htb]
    \centering
    \includegraphics[width=0.95\linewidth]{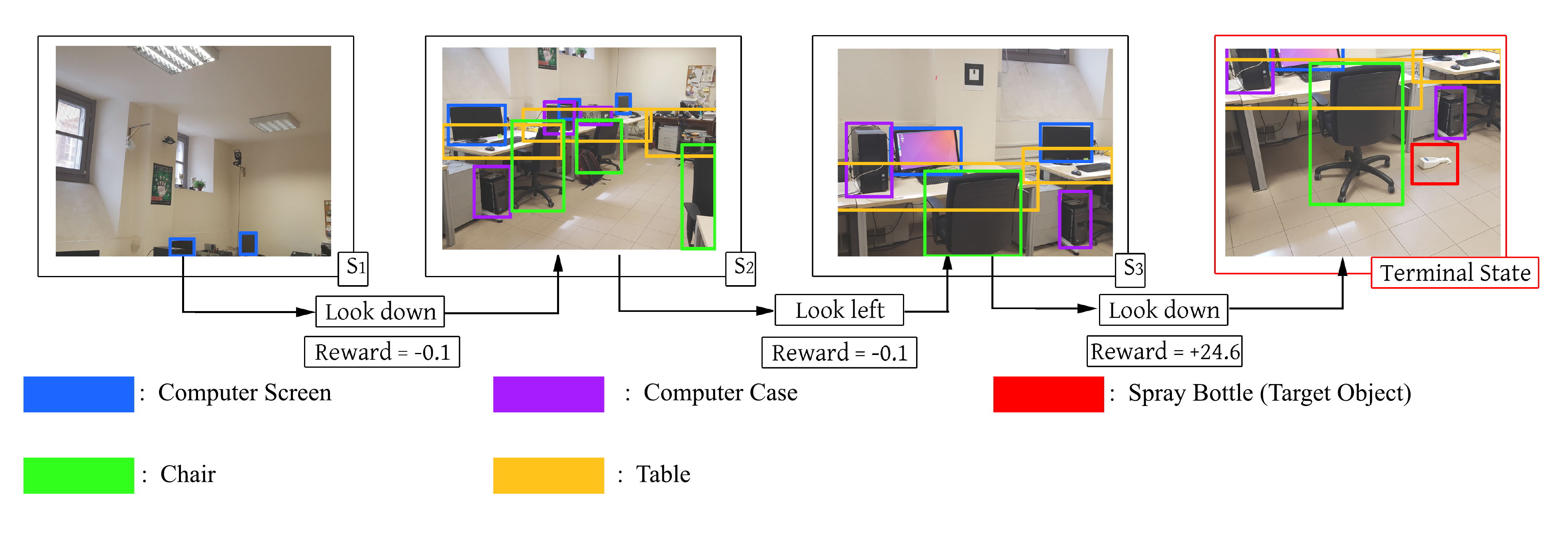}
    \caption{Successful Episode example of the Visual Search.}
    \label{fig:VisSearchSequence_1}
\end{figure}

\begin{figure}[!htb]
    \centering
    \includegraphics[width=0.95\linewidth]{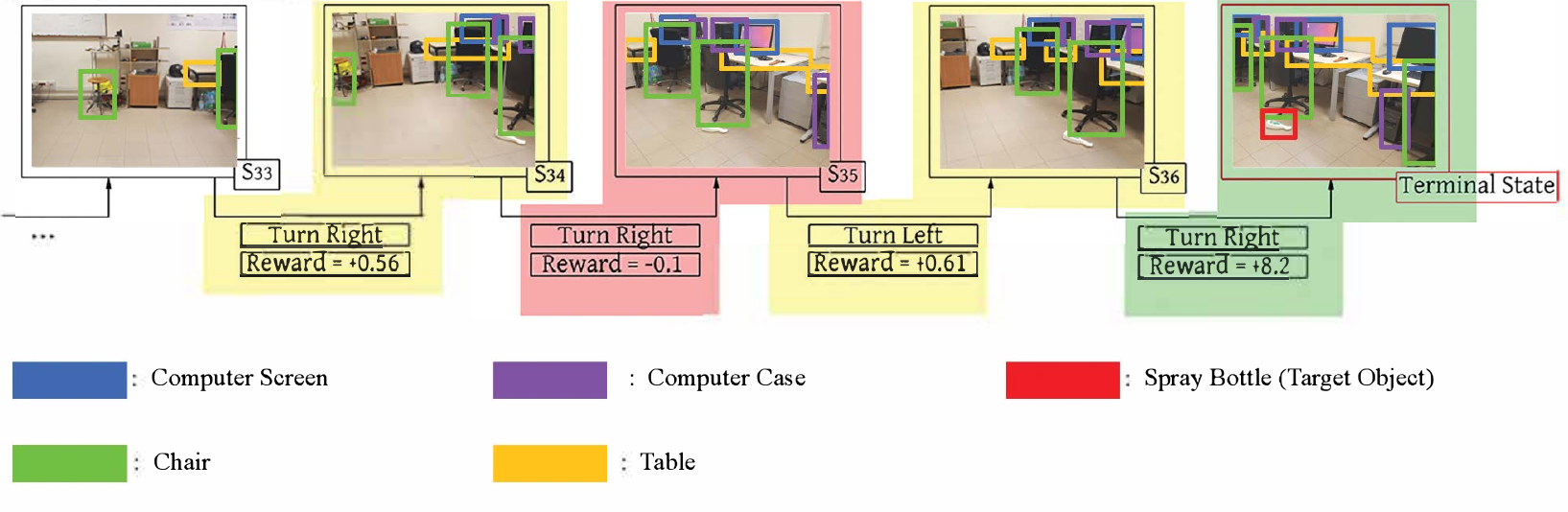}
    \caption{Object Detection failure in the Visual Search.}
    \label{fig:VisSearchSequence_2}
\end{figure}

\paragraph{\textbf{Evaluating results:}} An accurate evaluation of the robot  performances is a quite challenging task. There is no specific output value and the metrics that can be used tend to be very noisy.
To better evaluate the results of the training phase, the visual search policy has been tested in different ways. 100 thousand training steps have been done, at first by following a random policy and subsequently by following the policy returned from the trained network. In this setting, if the target is found in fewer than 50 steps the episode is ended and considered successful. In the case of random action choices, the success rate was of only 2.97\% with an average of 41 steps. With the trained network instead, the success rate has risen up to 88.47\% with an average of 19 steps as shown in Table \ref{VisSearchResults}.
\begin{table}[]
\centering
%\resizebox{0.99\columnwidth}{!}{
\begin{tabular}{|l|l|l|}
\hline
                                      & \textbf{Random Policy} & \textbf{Trained Policy} \\ \hline
\textbf{Successful Episodes}         & 59                     & 3919                    \\ \hline
\textbf{Unsuccessful Episodes}       & 1950                   & 510                     \\ \hline
\textbf{Total episodes in 100k Steps} & 1562                   & 4429                    \\ \hline
\textbf{Success Rate}                 & 2.97\%                 & 88.47\%                 \\ \hline
\textbf{Average Needed Success Step}         & 41                     & 19                      \\ \hline
\end{tabular}
%}
\caption{Visual Search test results}
\label{VisSearchResults}
\end{table}
To analyze the network behavior we tested the policy in the lab environment. The pictures returned from the camera and elaborated by the object detection network were collected together, with the action done and the reward received. Figures \ref{fig:VisSearchSequence_1} and Figure \ref{fig:VisSearchSequence_2} show some successful episode from the trained network. The trained network has learned an important principle: to look at the height at which the object is supposed to be found and then to look around. Figure \ref{fig:VisSearchSequence_2} also shows that sometimes the policy fails because the object detection network fails to correctly classify the target object.

\subsection{Experiments with the whole framework}

We evaluated two classes of tasks: (1) collect an object (spray bottle, screwdriver, hammer, cup) from the table or from the shelf to a subject;  (2) put-away an object (screwdriver, hammer, spanner) in the toolbox. Each experiment in a class has been run 35 times. An example is given in Figure \ref{fig:overview}, despite in a number of circumstances the grasping action has been manually helped, especially with small objects. 
Table \ref{tab:relations} shows the main relations, objects, and actions considered in the tasks. The table on the right side of Figure \ref{fig:histo1} shows the object detection accuracy achieved by the DCNN models dealing with the free and held objects during the above mentioned experiments.

%% Histos
\begin{figure}[t!]
 \centering
  \includegraphics[width=0.77\columnwidth]{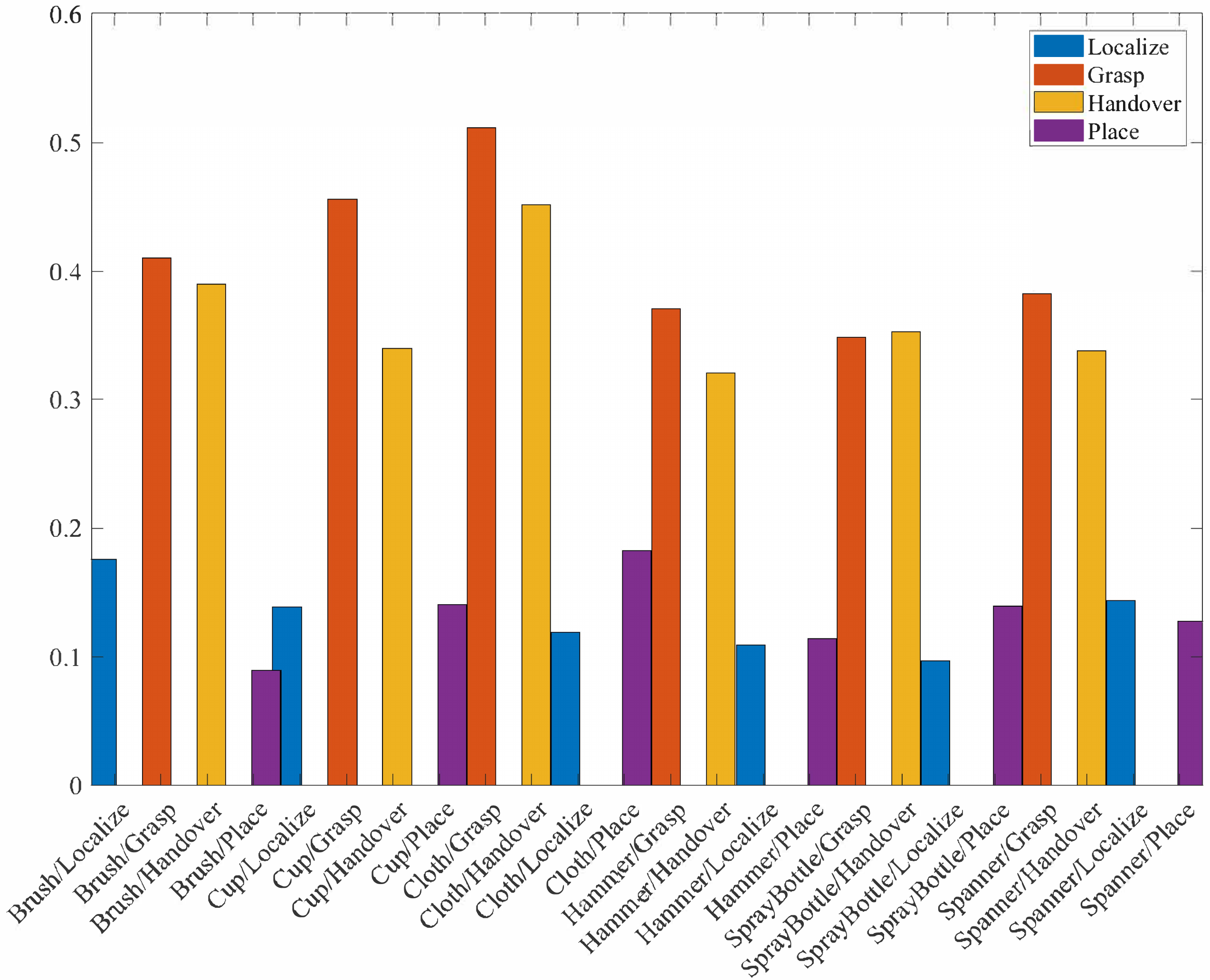}
 \caption{The histogram of action/objects failures. }\label{fig:histo2}
\end{figure}

\begin{figure}[t]
\includegraphics[width=0.77\columnwidth]{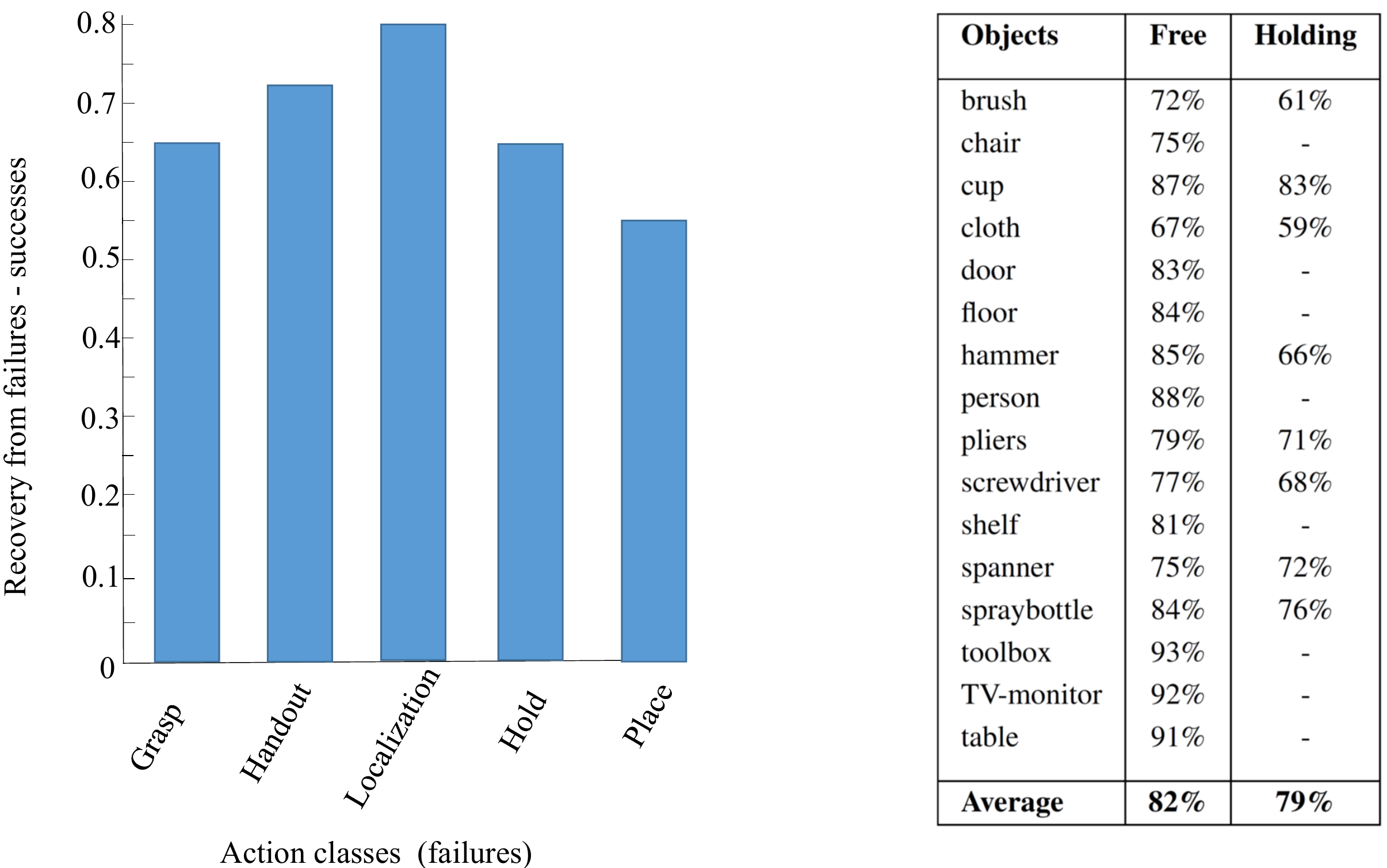}
 \caption{On the left, histogram of the recovery success ratios via the visual search policy. On the right, accuracy of the recognition of objects when handed by the robot or the human and when free. }\label{fig:histo1}
\end{figure}

%%%%%%%%%%%%%%%% FAILURES %%%%%%%%%%%%%%%%%%
\noindent
{\bf Failures}
We examined the number of failures encountered during the execution of the tasks described above. A failure is recorded as soon as the state perceived by the visual stream, via the DCNN,  does not match the post conditions of the action executed. The histogram in Figure \ref{fig:histo2} shows the probability that a failure is encountered while executing a particular action  in relation to the object being involved. We noticed that, as expected, more complex actions like grasping and localizing show a higher probability of failure. Surprisingly, passing an item shows a low failure probability, this is mainly attributed to the high adaptability of the subject involved in the action.

\noindent
{\bf Recovery}
The visual search policy is employed as soon as a failure is detected in order to find the primary objects involved in the action that failed. On the left side of Figure \ref{fig:histo1} it is shown the success rate of the recovery via the use of a visual search policy for four different types of actions. Localization and manipulation actions show a higher recovery success rate while recovering from a failed grasping seems to be the most problematic one.

\section{Conclusions}
In this work, we have proposed a new approach to execution monitoring by placing a strong emphasis on the role of  perception to both verify the state of execution and to recover in case of failure. Robot  representation  relies on  DCNNs to detect both objects and relations  and on Deep Reinforcement Learning to ensure awareness of the environment. 
The perception processes  are coordinated by the execution monitor, which uses them to either check the  realization of the properties required by the planner or to focus the robot attention onto the elements of the scene that are needed to complete the robot task.
We have shown that simple tasks can be completed in autonomy and that it is also possible to recover from failures. 
For future work, we are plan to explore new ways to complete the complex training phase for deep reinforcement learning, by directly learning from the environment in which the robot is operating

\section{Acknowledgments}
The research has been granted by the H2020 Project SecondHands under grant agreement No 643950.

%\bibliographystyle{IEEEtran}
%\bibliography{bibliography}


\end{document}